\pdfoutput=1

\documentclass[11pt]{article}

\usepackage{EMNLP2023}

\usepackage{times}
\usepackage{latexsym}
\usepackage{makecell}

\usepackage[T1]{fontenc}

\usepackage[utf8]{inputenc}

\usepackage{microtype}

\usepackage{inconsolata}

\usepackage{booktabs}
\usepackage{multirow}
\usepackage{xcolor}
\usepackage{tabularx}
\usepackage{tikz}
\usepackage{amsmath}
\usepackage{hyperref}
\usepackage{amssymb}%
\usepackage{pifont}%
\usepackage{subcaption}
\usepackage{graphicx} %
\usepackage{array} %
\usepackage{cellspace} %
\usepackage{fontawesome5} %
\usepackage{soul}
\usepackage{paralist} %

\usepackage{gb4e}
\noautomath

\def\Snospace~{\S{}}
\renewcommand\sectionautorefname{\Snospace}

\usepackage{etoolbox}
\AtBeginEnvironment{quote}{\itshape}
\AtBeginEnvironment{nquote}{\itshape}

\makeatletter
\renewcommand{\sectionautorefname}{\S\@gobble}
\renewcommand{\sectionautorefname}{\S\@gobble}
\renewcommand{\subsectionautorefname}{\S\@gobble}
\renewcommand{\sectionautorefname}{\S\@gobble}
\renewcommand{\subsectionautorefname}{\S\@gobble}
\renewcommand{\appendixautorefname}{\S\@gobble}
\makeatother

\DeclareSymbolFont{extraup}{U}{zavm}{m}{n}
\DeclareMathSymbol{\vardiamond}{\mathalpha}{extraup}{87}
\newcommand{\aspace}{\hspace{1em}}
\newcommand{\uw}{$^{\heartsuit}$}
\newcommand{\nyu}{$^{\spadesuit}$}
\newcommand{\aiTwo}{$^{\clubsuit}$}

\title{How Language Model Hallucinations Can Snowball}

\author{Muru Zhang\uw \aspace Ofir Press\uw  \quad William Merrill\nyu  \quad Alisa Liu\uw  \aspace Noah A.~Smith\uw\aiTwo\\
  \uw Paul G.~Allen School of Computer Science and Engineering, University of Washington \\ 
  \nyu New York University \\
  \aiTwo Allen Institute for Artificial Intelligence \\ 
  \texttt{nanami17@cs.washington.edu} \\ }

\begin{document}
\maketitle
\begin{abstract}
A major risk of using language models in practical applications is their tendency to hallucinate incorrect statements. %
Hallucinations are often attributed to knowledge gaps in LMs, but we hypothesize that in some cases, when justifying previously generated hallucinations, LMs output false claims that they can separately \textit{recognize} as incorrect.
We construct three question-answering datasets where ChatGPT and GPT-4 often state an incorrect answer and offer an explanation with at least one incorrect claim. 
Crucially, we find that ChatGPT and GPT-4 can identify 67\% and 87\% of their own mistakes, respectively.
We refer to this phenomenon as hallucination snowballing: an LM over-commits to early mistakes, leading to more mistakes that it otherwise would not make.\footnote{Our data and code are available at: \url{https://github.com/Nanami18/Snowballed_Hallucination}}
\end{abstract}
\section{Introduction}

\begin{table*}[h]
    \centering
    \resizebox{\textwidth}{!}{%
    \begin{tabular}{cll}
        \toprule
        \textbf{Dataset} &\textbf{Original Question} & \textbf{Verification Question}\\
        \midrule
        \makecell[c]{\calculator{}\\ \textbf{Primality}\\ \textbf{Testing}} &
        \makecell*[{{p{10cm}}}]{
        \faUser~~\textbf{User:} Is 10733 a prime number?\\
        \faRobot~\textbf{GPT-4:} No... It can be \underline{factored into 3 $\times$ 3577}.} 
        & \makecell*[{{p{6cm}}}]{
        \faUser~~\textbf{User:} Is 10733 divisible by 3? Answer with either Yes or No.\\
        \faRobot~ \textbf{GPT-4:} \underline{No}
        }\\
        \midrule 
        \makecell[c]{\senator{}\\ \textbf{Senator}\\ \textbf{Search}}&\makecell*[{{p{10cm}}}]{
        \faUser~~\textbf{User:} Was there ever a US senator that represented the state of New Hampshire and whose alma mater was the University of Pennsylvania?\\
        \faRobot~ \textbf{GPT-4:} Yes... His name was \underline{John P. Hale}} 
        & \makecell*[{{p{6cm}}}]{
        \faUser~~\textbf{User:} Was John P. Hale's alma mater University of Pennsylvania?\\
        \faRobot~ \textbf{GPT-4:} \underline{No}... [it] was Bowdoin
        }\\ %
        \midrule
        \makecell[c]{\graph{}\\\textbf{Graph}\\\textbf{Connectivity}}&\makecell*[{{p{10cm}}}]{\faUser~~\textbf{User:} Current flight information (the following flights are one-way only, and all the flights available are included below):\\
        There is a flight from city F to city K\\
        There is a flight from city H to city A\\
        \textit{[... 10 other rules cut for space ...]}\\
        Question: Is there a series of flights that goes from city B to city E?\\
        \faRobot~ \textbf{GPT-4:} Yes... the route is as follows: ... \underline{City K to City G}...
        } 
        & \makecell*[{{p{6cm}}}]{
        \faUser~~\textbf{User:} \textit{[...flight information given in the context...]} \\Based on the above flight information, is City K to City G a valid flight?\\
        \faRobot~\textbf{GPT-4:} \underline{No}, based on the above flight information, there is no direct flight from City K to City G.
        }\\
        \bottomrule 
    \end{tabular}}
    \caption{\textbf{Original Question}: An example from each dataset and GPT-4's (incorrect) response, with the snowballed hallucination \underline{underlined}. \textbf{Verification Question}: The corresponding query used to probe whether GPT-4 recognizes its claim is incorrect, with its refutation of the hallucination \underline{underlined}.}
    \label{tab:task example}
\end{table*}
Language models are increasingly being deployed to interface with humans in open-ended information-seeking and problem-solving settings.
Despite their diverse capabilities and extreme fluency, a major open challenge is that LMs still hallucinate by making up facts or citing sources that do not exist~ \cite[i.a.]{maynez-etal-2020-faithfulness, liu2023evaluating}, often while sounding extremely plausible.

\begin{figure}[h!]
    \centering
    \includegraphics[width=\columnwidth]{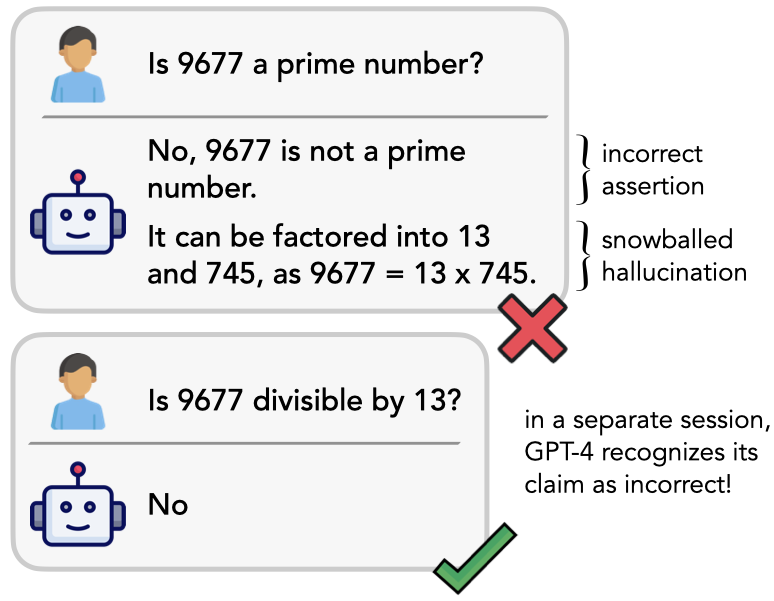}
    \caption{GPT-4 mistakenly claims that 9677 is not prime, followed by an incorrect explanation that 13 $\times$ 745 $=$ 9677. 
    We refer to this factorization as a \textit{snowballed hallucination}, as GPT-4 appears to ``know'' that 13 is not a factor of 9677 when asked separately.
    Thus, hallucinations are not necessarily reflective of a model's lack of ``knowledge,'' but may be induced by an earlier hallucination for consistency.}
    \label{fig:figure1}
\end{figure}

Hallucination is commonly attributed to knowledge gaps in LMs \cite{zheng2023does}, motivating mitigation strategies through retrieval over knowledge bases \cite{lewis-etal-2020-retrieval,shuster-etal-2021-retrieval-augmentation, peng2023check}
But, do LMs \emph{only} hallucinate when they do not ``know'' a fact?  %
We present a setting where LMs often generate hallucinations that they immediately %
recognize as wrong when presented in isolation. 
Specifically, after an LM answers a question incorrectly, it usually justifies that answer by making incorrect assertions that it separately acknowledges as incorrect (\autoref{fig:figure1}). %

To study this behavior empirically, we automatically construct three question-answering (QA) datasets.
These datasets span different domains: determining whether a number is prime, whether there is a U.S.~senator satisfying two given constraints, and whether two cities are connected given a set of flights between cities.
Empirically, we find that ChatGPT \cite{openai-2022-chatgpt} and GPT-4  \cite{gpt4} commit to an answer within the first token (\texttt{Yes}/\texttt{No}) over 95\% of the time;
these answers are often incorrect, and then followed by an incorrect explanation.
Yet, when presented with the incorrect explanation alone, we find that the LM is likely able to recognize it as incorrect.

We refer to this phenomenon as \textbf{hallucination snowballing}.
We hypothesize that LMs produce snowballed hallucinations for consistency with earlier hallucinations (rather than due to a ``knowledge gap'' in the model), as they recognize the snowballed hallucination is incorrect when presented in isolation (i.e., in a separate interaction session).

While prompting strategies that encourage the LM to reason before stating an answer improve accuracy on the task, our work points to the broader issue that conditioning on faulty context leads LMs to produce extremely simple mistakes that they wouldn't otherwise make.
Indeed, when prompting with ``\textit{Let's think step by step}'' \cite{kojima2023large},
snowballed hallucinations still occur in 95\% of cases where the model fails to answer correctly. 
We observe that sometimes even when ``\textit{Let's think step by step}'' does lead to the right answer, it uses invalid reasoning chains.

In this paper, we demonstrate the phenomenon of hallucination snowballing by leveraging recent LMs' tendency to state and justify their answers.
Rather than over-committing to its previously generated context, %
we believe that LMs should acknowledge their initial mistake, and then revise their answer.
We have indeed observed GPT-4 doing this in a limited number of cases;  amplifying this behavior would be beneficial, as well as developing new methods in which LMs can backtrack.

\section{Why do we expect hallucination snowballing?} \label{sec:why}
In this section, we explain why we hypothesize that LMs are susceptible to hallucination snowballing. We predict that snowballing will occur on questions with two key properties:
\begin{compactenum}
    \item \textbf{Initial committal:} The prompt leads the LM to first state an answer (\emph{before} outputting the explanation). This applies to many yes/no questions.%
    \item \textbf{Inherently sequential:} Transformers cannot find the answer within one timestep because of their limited reasoning abilities within one timestep. 
\end{compactenum}
We now discuss how these properties may lead to snowballed hallucination.%

\paragraph{Initial committal.} In English and many other languages, speakers often say the final Yes/No answers to questions before explaining their answer. 
We therefore hypothesize that LMs and especially instruction-tuned LMs~\cite{wei2021finetuned,sanh2021multitask,instructGPT,wang2022selfinstruct} will reflect this answer format where the answer comes before the explanation.
Indeed, on our datasets (presented in \S\ref{sec:datasets}), we observe that GPT-4 and ChatGPT immediately commit to an answer to the question: the first token is \texttt{Yes} or \texttt{No} 95.67\% and 98.40\% of the time for GPT-4 and ChatGPT respectively. %
In the remaining cases, the model often commits to an answer within the first few tokens of the response (e.g., \textit{``There is no record of a U.S. Senator...''}).
Crucially, once the LM generates \texttt{Yes} or \texttt{No}, that token remains in the context, and coherence would require commitment to that choice through the subsequent justification. Thus, the model produces an answer to a complex question in a \emph{single} timestep, and it then continues by generating an explanation for that answer, which inevitably will be incorrect.

\paragraph{Inherently sequential.}
Furthermore, transformers cannot solve inherently sequential reasoning problems like primality testing or graph connectivity within a single timestep,\footnote{Technically, this holds only for inputs above a certain hardness level, i.e., the size of the prime number for primality testing, or the size of the graph for graph connectivity.} as documented in recent theoretical results \citep{merrill2023parallelism}.\footnote{\citet{merrill2023parallelism} show that, with a single generation step, bounded-precision transformers cannot solve any problem outside the complexity class $\mathsf{TC}^0$, which corresponds to a highly parallelizable subclass of both $\mathsf{L}$ (log-space) and $\mathsf{P}$ (polynomial-time). Graph connectivity is an $\mathsf{L}$-complete problem, which means it cannot be in $\mathsf{TC}^0$ unless $\mathsf{TC}^0 = \mathsf L$, i.e., all of $\mathsf L$ can be parallelized to a surprisingly high degree. Primality testing was shown to be in $\mathsf P$ \citep{agrawal2004primes} but cannot be in $\mathsf{TC}^0$ unless it is also in $\mathsf L$; i.e., any $n$ can be factored with $O(\log \log n)$ bits of overhead.
In summary, unless standard complexity-theoretic conjectures are false, graph connectivity and primality testing are outside $\mathsf{TC}^0$ and thus are too inherentially sequential for transformers to solve in a single generation \citep[cf.][]{merrill2023parallelism}.}
Our graph connectivity and primality datasets are concrete instantiations of these problems. Because the transformer must use one step to answer a question that requires multiple timesteps to answer correctly, it will necessarily sometimes commit to an incorrect answer. We hypothesize that this leads the LM to hallucinate supporting incorrect facts that it otherwise would not generate.%

\section{Experiments} \label{sec:exp}

We design three QA datasets with the properties described in \autoref{sec:why} to probe hallucination snowballing, and evaluate ChatGPT and GPT-4. We first check whether the LM returns the correct answer to the given question, and we show that when the model returns the wrong answer, it frequently provides an incorrect explanation for that wrong answer. We automatically extract the incorrect claim in the explanation and ask the same LM to check whether its claim is correct. 
See \autoref{tab:task example} for a representative example from each dataset.

\subsection{Datasets} \label{sec:datasets}
We design three QA datasets, 
each containing 500 yes/no questions 
that we expect are not answerable by transformers in one timestep. %
To aid evaluation, the questions are designed so that an incorrect answer would be justified with easily verifiable claims. %

For each dataset, we fix one specific label for all examples, so that if the model chooses the incorrect answer (e.g., that 9677 is not prime), it would produce a specific claim to support it (e.g., an incorrect factorization). This enables us to systematically examine model-written justifications for incorrect answers.

\paragraph{\calculator{} Primality testing}
For this dataset, we query the primality of 500 randomly chosen primes between 1,000 and 20,000; the correct answer is always \texttt{Yes}.
When the model answers incorrectly, we expect it to justify its answer with an incorrect factorization.

\paragraph{\senator{}~~Senator search}
This dataset consists of 500 questions of the form ``\textit{Was there ever a US senator that represented the state of $x$ and whose alma mater was $y$?}'' where $x$ is a U.S. state and $y$ is a U.S. college. For these questions, the correct answer is always \texttt{No}. When the model answers incorrectly, we expect it to falsely claim that a particular senator both represented $x$ and attended $y$.

To create the dataset we consider all U.S. states and a manually constructed list of twelve popular U.S. colleges (see \autoref{sec:appendix-dataset} for the full list);
for each possible pair, we generate a question following the template,
and manually remove pairs where the answer is \texttt{Yes}. 

\begin{figure*}[]
    \centering
    \begin{subfigure}{\textwidth}
        \centering
        \medskip
        \includegraphics[width=.45\linewidth]{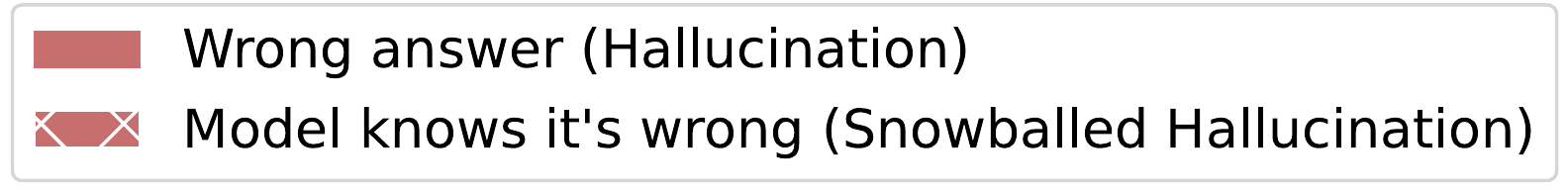}
    \end{subfigure}
    \begin{subfigure}{.45\textwidth}
        \centering
        \includegraphics[width=\linewidth]{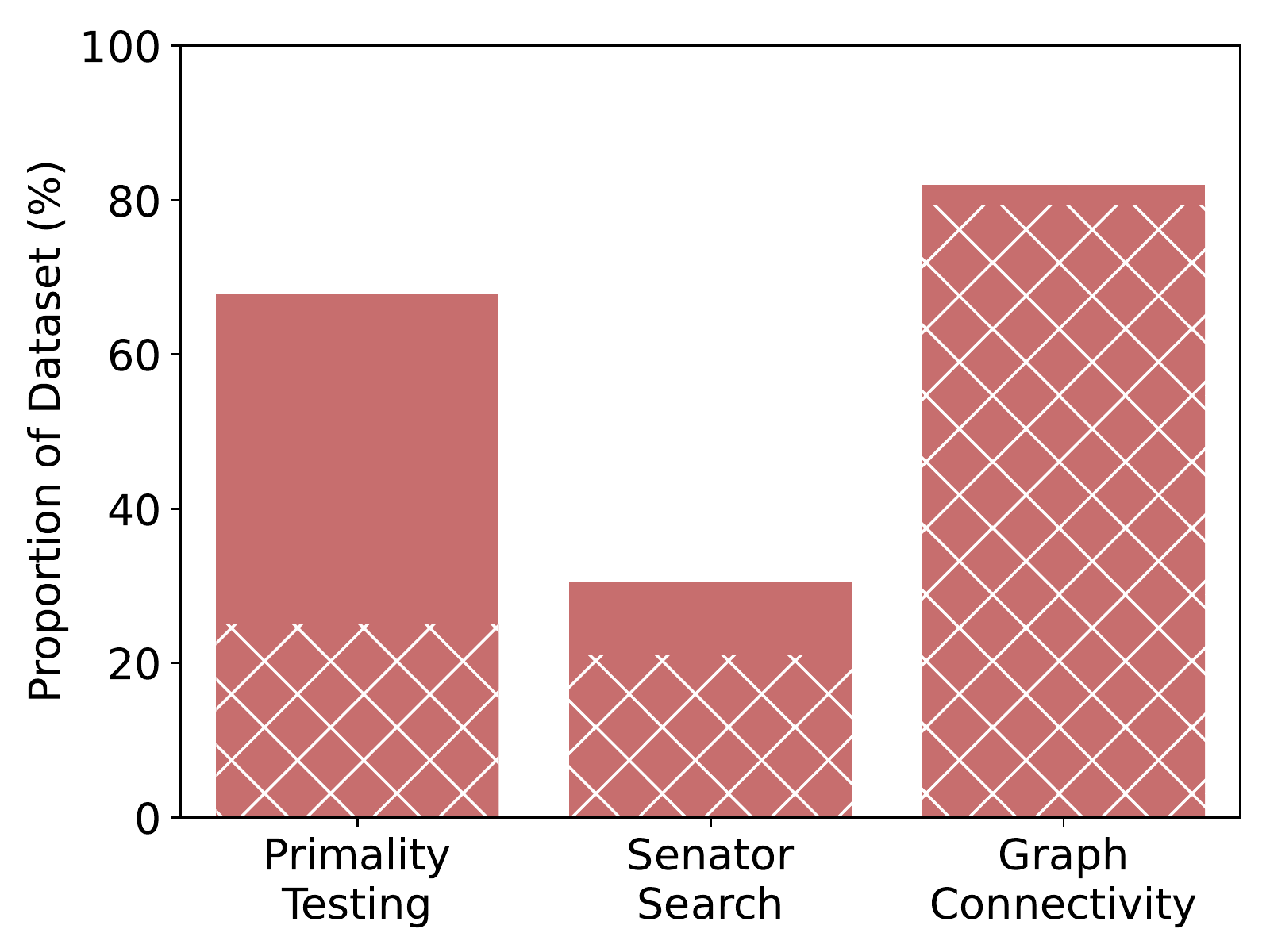}
        \caption{ChatGPT}
    \end{subfigure}%
    \begin{subfigure}{.45\textwidth}
      \centering
      \includegraphics[width=\linewidth]{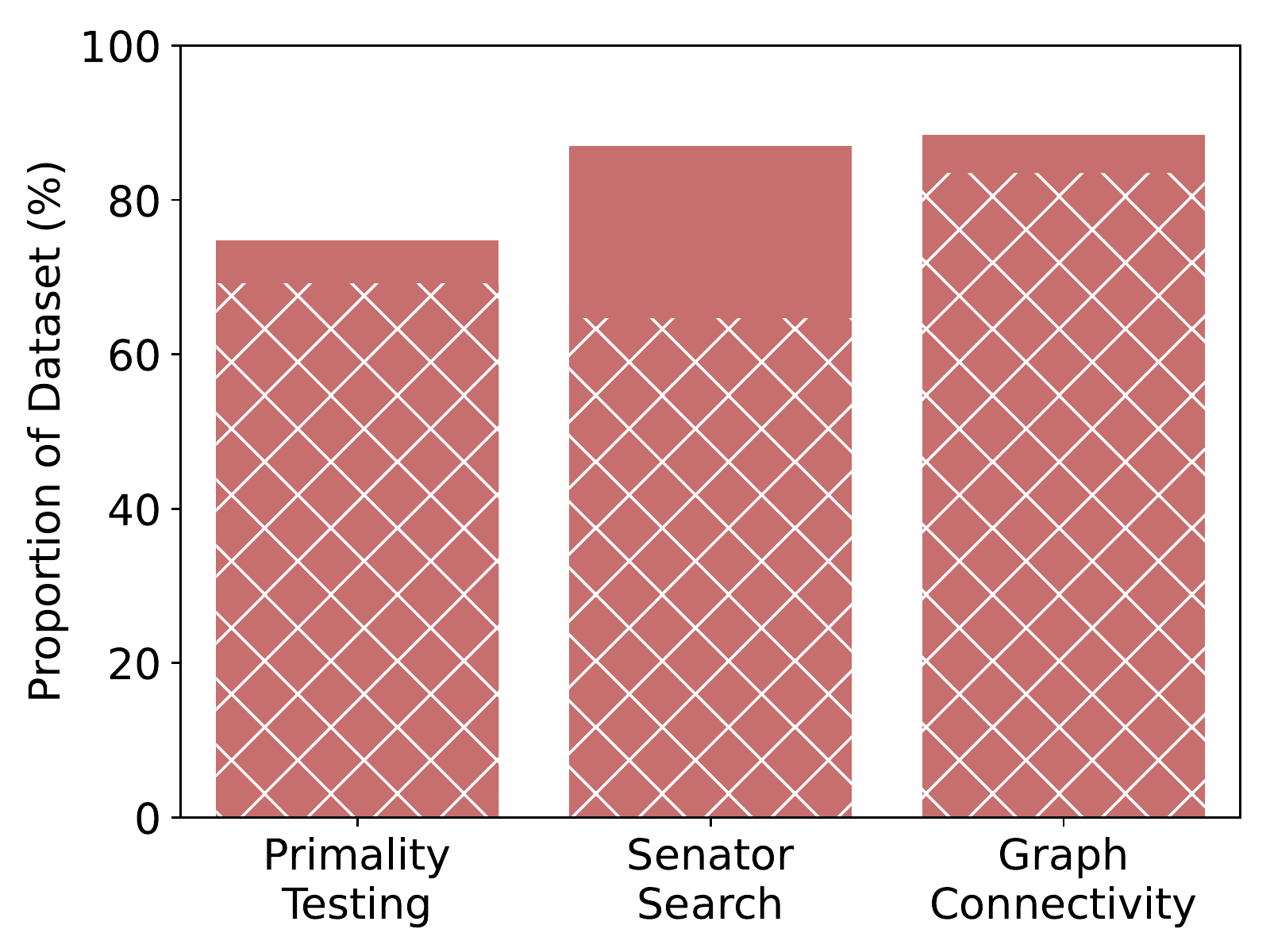}
      \caption{GPT-4}
    \end{subfigure}
    \caption{Percentage of hallucination and percentage of snowballed hallucination (both calculated with respect to the entire dataset) for ChatGPT and GPT-4. The precise numbers for this plot are available in \autoref{tab:task performance} and~\autoref{tab:verification performance} in the Appendix.}
    \label{fig:hallu and forced hallu plot}
\end{figure*}

\paragraph{\graph{} Graph connectivity}
For each of the 500 questions in this dataset, we present 12 flights among 14 cities, and ask if there is a sequence of flights from a particular city to another.
The problem always corresponds to the same underlying directed graph structure (see \S\ref{subsec:appendix-graph-connectivity}), where flights are edges and cities are nodes.
For each instance in the dataset, we randomly assign letters from the English alphabet to name the nodes.
To formulate the query, we sample a source city $s$ and destination city $t$ in different subgraphs, with the additional constraint that $s$ corresponds to a source node, and $t$ a leaf node, so that 1-step heuristics cannot be used to solve the problem. 

We formulate the problem as a flight-finding question in natural language so that it sounds more natural: in the prompt, we list the twelve flights (``\textit{There is a flight from city F to city K; there is a flight from city G to city N,} ...''), followed by the question \textit{``Is there a series of flights... from $s$ to $t$?''}.
Note the correct answer is always \texttt{No}.
When the model answers incorrectly, we expect it to justify its answer with a flight that does not exist.

\subsection{Inference Setup}
\textbf{Language models.} We run all experiments on ChatGPT (\texttt{gpt-3.5-turbo}) and GPT-4 with greedy decoding. 

Our experiments are \textit{zero-shot} (i.e., we do not show the model any example QA pairs in the prompt). 
We focus on the model behavior under the direct prompt (see \autoref{sec:appendix-dataset} for full examples), which is the most common way users interact with LMs. See \autoref{sec:how to fix} for experiments with the zero-shot chain-of-thought style prompting method.

For each dataset, we perform a two-stage evaluation. %
First, we evaluate the model's accuracy (i.e., how many of the questions it answers correctly). 
When either models is \emph{incorrect}, empirically it \emph{always} generates a justification. In the second stage, we assess whether the model can identify the 
incorrect step in the explanation. 

For a given question, we evaluate the model's response by examining whether the output begins with either \texttt{Yes} or \texttt{No}. In cases where the response does not fall into these categories, we manually determine the answer conveyed by the model.

\subsection{LM Recognition of Snowballed Hallucinations} 
We probe whether LMs recognize their snowballed hallucinations by verifying the model's incorrect claims in the output against the model itself. Note that our recognition procedure relies on heuristics gained from manual examination of the model output, and these heuristics might not work on other models (e.g., a different model might not provide factors when supporting the claim that a number is not prime).

\paragraph{Graph Connectivity}
For each sample where the model thinks there is a series of connecting flights (where answer starts with \texttt{Yes}), we manually extract the list of flights from the model's output and identify the invalid or discontinuous flights. 

We then, in a new session, ask the model to verify whether the extracted flights are valid based on the flight information, and if consecutive flights are indeed connected. We manually assess the verification output to check if the model correctly detects the error. See Appendix \autoref{tab:GC verification example} for how we prompt the model and an example of successful verification.

\paragraph{Primality Testing}
For each sample where the model answers that the number is not prime, we extract the factors the model uses to justify it. The extraction is done by putting the output in the context and asking \textit{``What are the factors proposed in the above text? List them out.''}
We use ChatGPT for extraction with one-shot demonstration (for its fast inference speed); we manually checked 30 examples and found that it can always extract the correct factors.

We then, in a new session, ask the model to verify each extracted factor individually. See Appendix \autoref{tab:PT verification example} for an example of successful verification.

\begin{figure*}[t!]
\centering
    \begin{subfigure}{\textwidth}
        \centering
        \medskip
        \includegraphics[width=.4\linewidth]{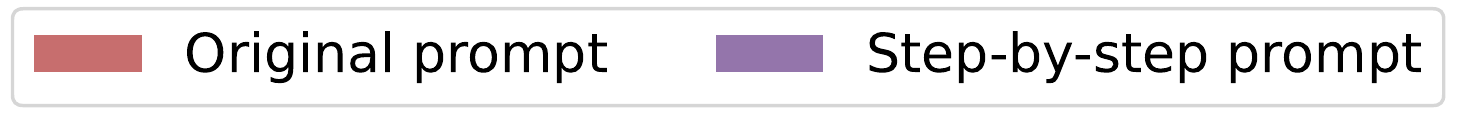}
    \end{subfigure}
    \begin{subfigure}{.45\textwidth}
        \centering
        \includegraphics[width=\linewidth]{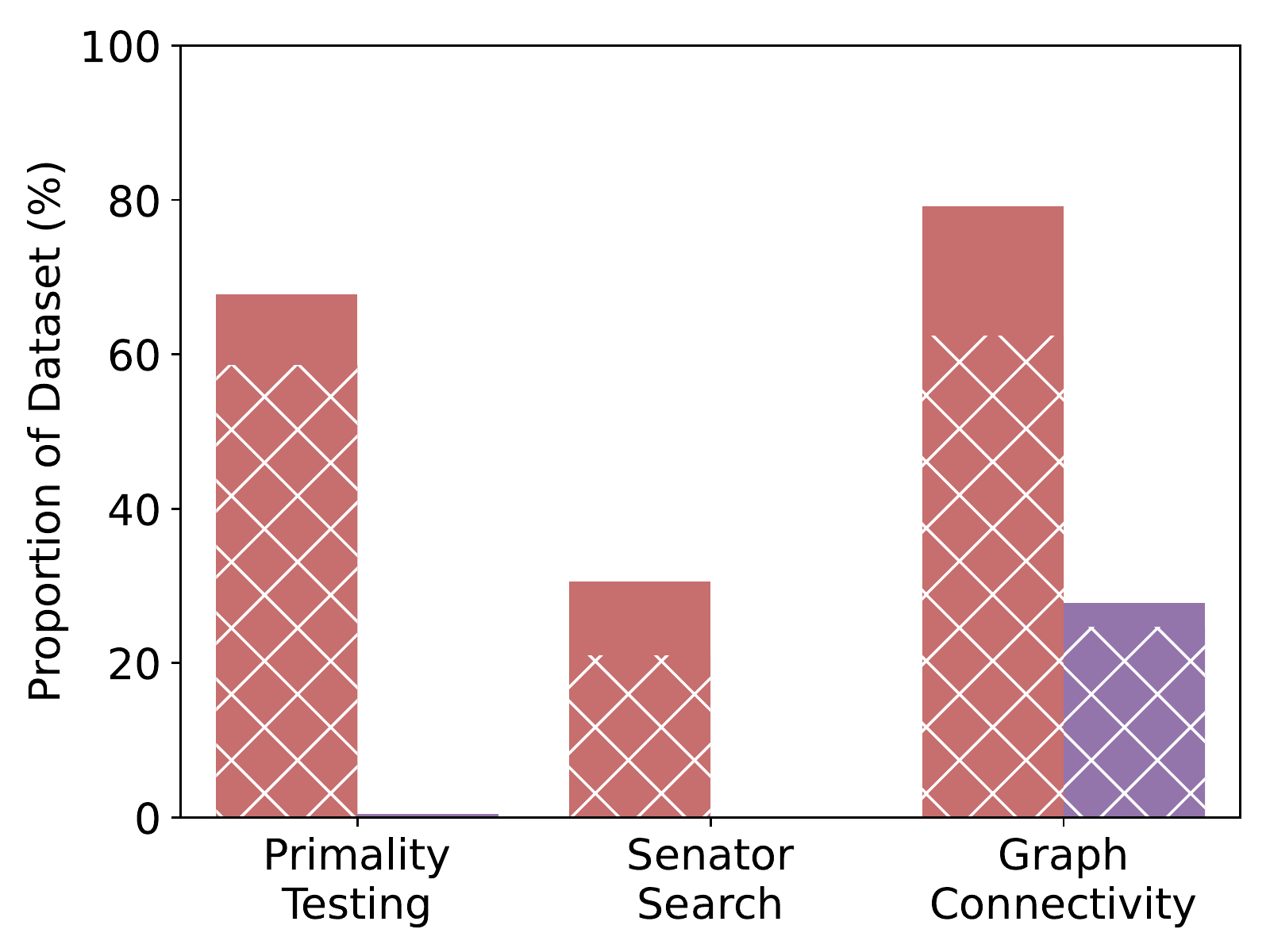}
        \caption{ChatGPT}
    \end{subfigure}%
    \begin{subfigure}{.45\textwidth}
        \centering
        \includegraphics[width=\linewidth]{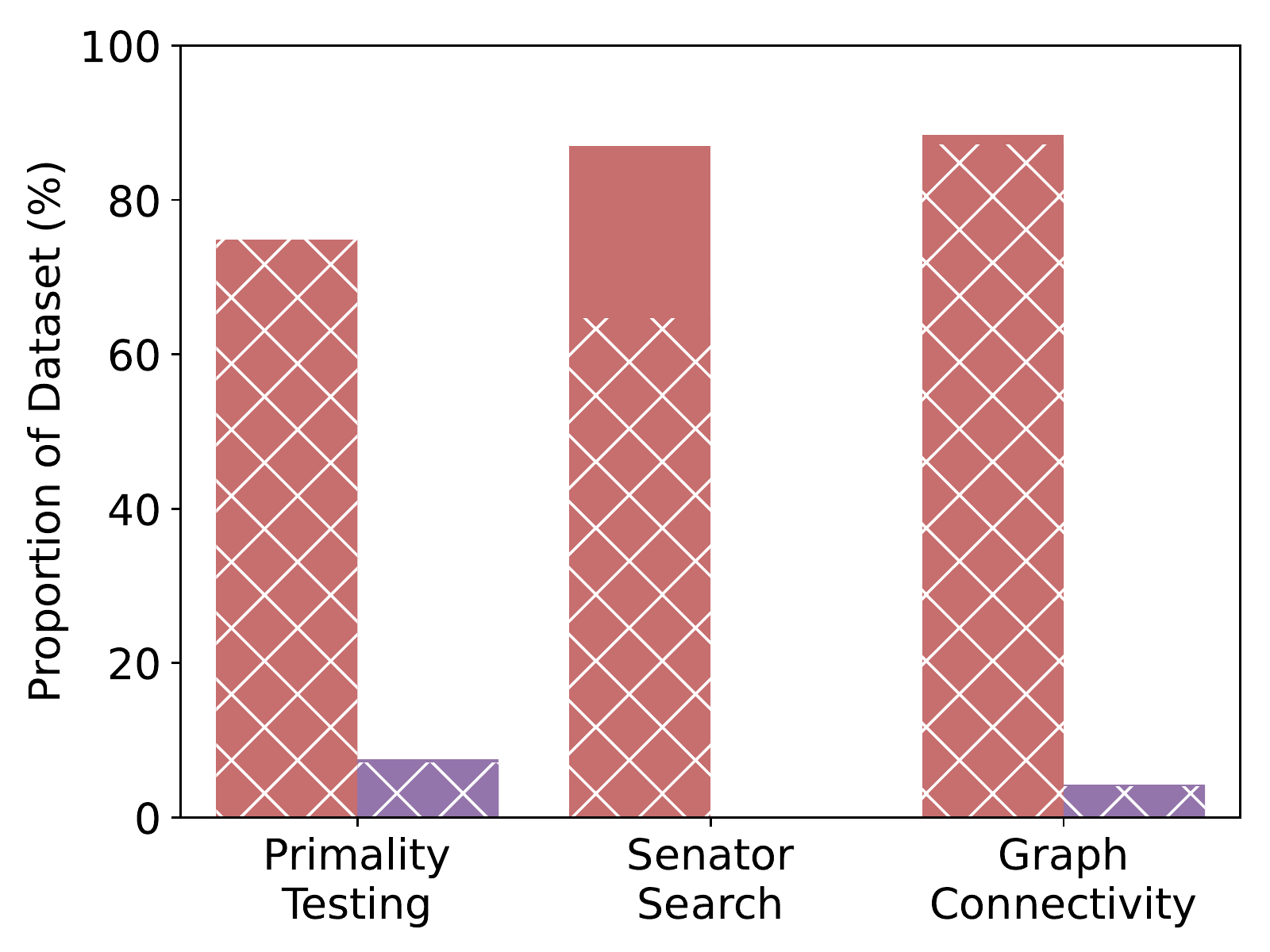}
        \caption{GPT-4}
    \end{subfigure}
\caption{Error rate and snowballed hallucination rate (hatch pattern) for ChatGPT and GPT-4, when using the original prompt versus \textit{``Let's think step by step''}. See Appendix \autoref{tab:lts task performance} and \autoref{tab:lts verification performance} for the exact numbers.} 
\label{fig:cot_plot}
\end{figure*}

\paragraph{Senator Search}
For each sample where the model thinks there is such senator, we extract the name of the senator the model uses to justify the existence, by putting the output in the context and asking \textit{``What is the senator mentioned in the above text? Just give the name''}. Again, we use ChatGPT and manually observed perfect extraction on 30 examples.

We then, in a new session, ask the model if that senator's alma mater is the college in the question and has represented the state in the question. See Appendix \autoref{tab:SA verification example} for an example of successful detection.

\subsection{Results}
\paragraph{Question-answering accuracy}
\autoref{fig:hallu and forced hallu plot} shows that both ChatGPT and GPT-4 experience very low accuracy across the board. With the exception of ChatGPT on the \textbf{Senator Search} dataset,
all models achieve less than 50\% accuracy.%
(See Appendix \autoref{tab:task performance} for a breakdown of the error rate by dataset.)
We observe that GPT-4 performs worse than ChatGPT across all datasets 
despite popularly being considered superior to ChatGPT~\cite{gpt4}.  %
While ChatGPT has an average accuracy of $39.87\%$, GPT-4 has only $16.6\%$.

\paragraph{Hallucination detection}

Here, we check whether the model can identify that the incorrect claim is wrong when it is presented alone.
As shown in \autoref{fig:hallu and forced hallu plot}, ChatGPT detects 67.37\% 
of incorrect claims in explanations (i.e., snowballed hallucinations), and GPT-4 detects 87.03\%.
Notice that when the model fails the verification (an example in Appendix \autoref{tab:failed verification}), 
we do not consider it a snowballed hallucination.%

Overall, we find that ChatGPT and GPT-4 are both extremely susceptible to hallucination snowballing, leading to extremely simple mistakes.

\begin{figure*}[t!]
\centering
    \begin{subfigure}{\textwidth}
        \centering
        \includegraphics[width=0.3\linewidth]{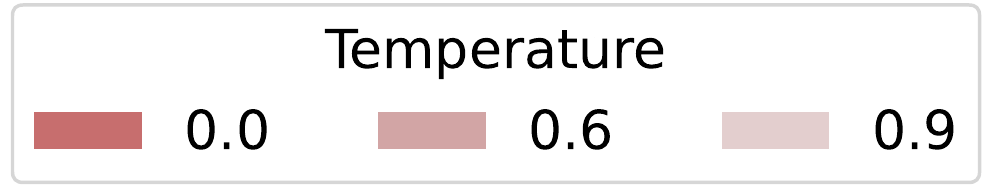}
    \end{subfigure}
    \begin{subfigure}{.45\textwidth}
        \centering
        \includegraphics[width=\linewidth]{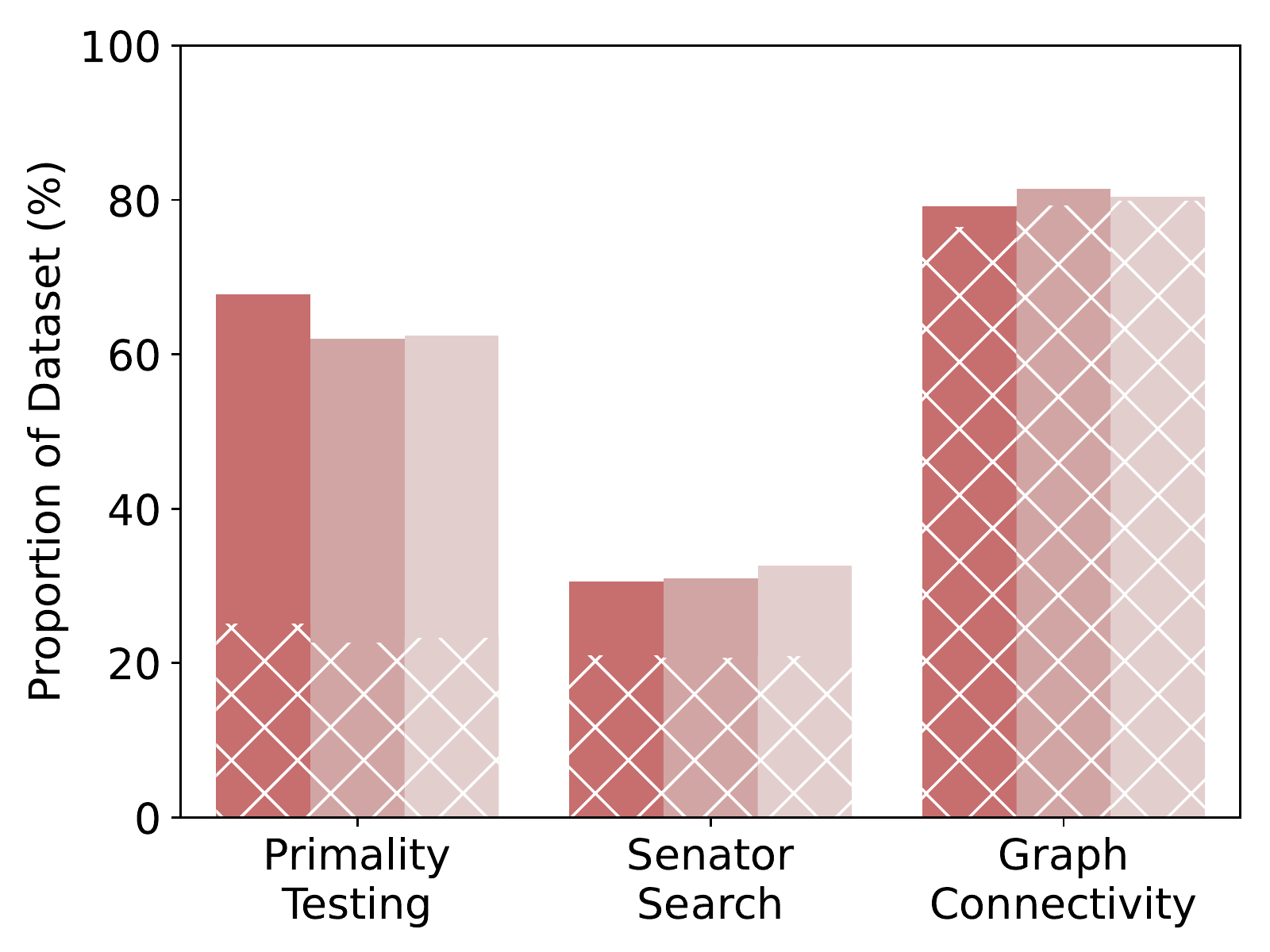}
        \caption{ChatGPT}
    \end{subfigure}%
    \begin{subfigure}{.45\textwidth}
        \centering
        \includegraphics[width=\linewidth]{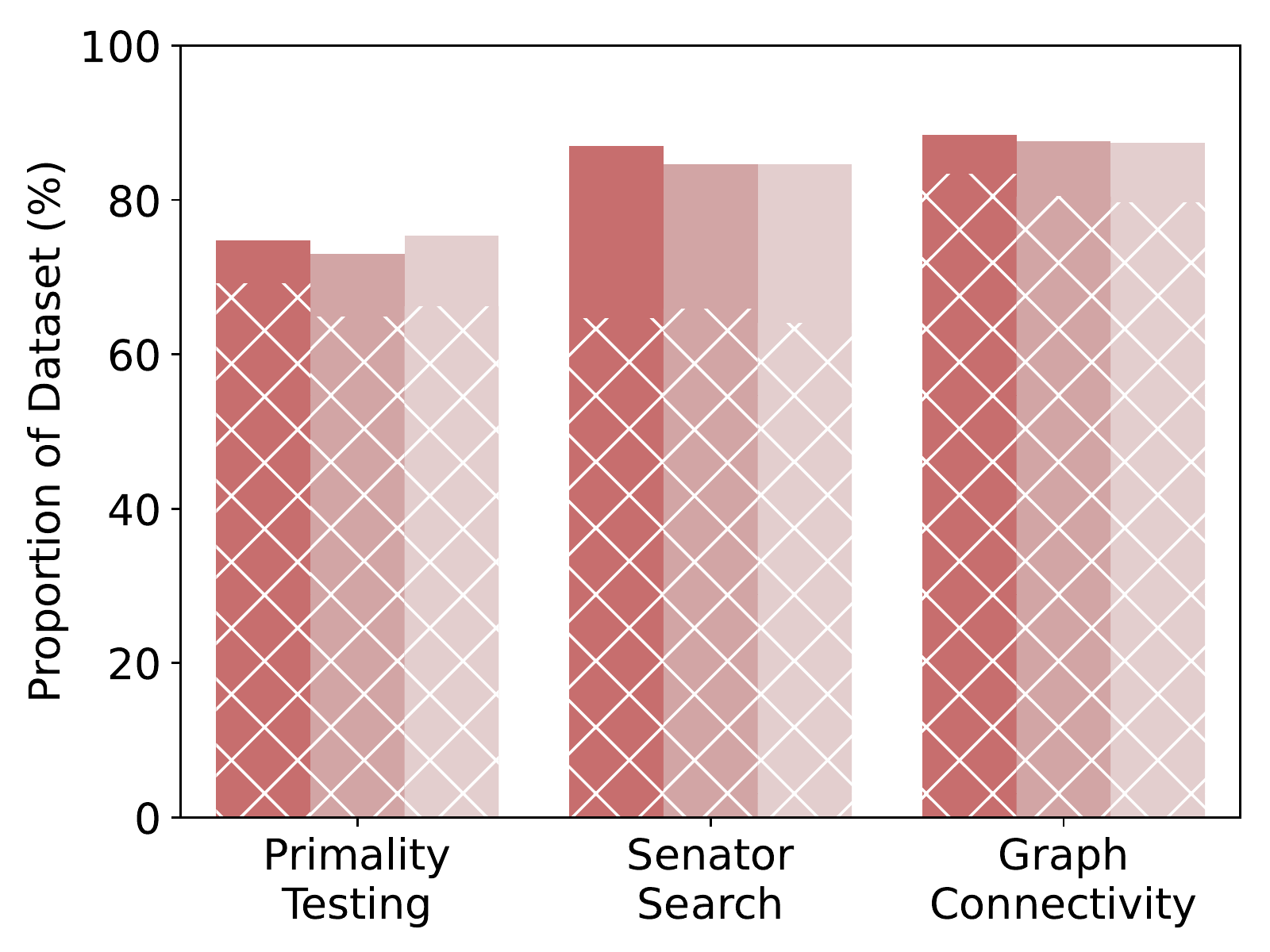}
        \caption{GPT-4}
    \end{subfigure}
    \caption{Error rate and snowballed hallucination rate (hatch pattern) from ChatGPT and GPT-4, when using different values for temperature at decoding-time. See Appendix \autoref{tab:temperature experiment - accuracy} and \autoref{tab:temperature experiment - snowball} for the exact numbers.} 
    \label{fig:temperature_plot}
\end{figure*}

\section{Can we prevent snowball hallucinations?} \label{sec:how to fix}
We hypothesize that hallucination snowballing occurs because LMs are trained to model continuations consistent with their current context (the given prompt and prior outputs). Although a fix to the fundamental problem might require more than just inference-time modification, in this section we study the effectiveness of two inference strategies in alleviating hallucination snowballing: prompting (\autoref{subsec:better_prmopts}) and decoding or training methods (\autoref{subsec:algorithmic_corrections}).

\subsection{Engineering Better Prompts}\label{subsec:better_prmopts}

In this section, we examine the effectiveness of better prompts on preventing snowballed hallucination by using a different zero-shot prompt that encourages the model to generate the reasoning chain before the answer.
Since the outputs generated under these prompts are less structured, we manually inspect them to determine correctness and the presence of snowballed hallucinations.

For each task, we append \textit{``Let's think step-by-step''} at the end of the original question (shown in \autoref{tab:task example}). As shown in \autoref{fig:cot_plot}, the model can solve the \textbf{Senator Search} task perfectly, achieve $\leq$10\% error rate on \textbf{Primality Testing}, and $\leq$30\% on \textbf{Graph Connectivity}. 
Despite the large improvement in accuracy, we identify a potential issue: %
the model sometimes  hallucinate while outputting the reasoning chain, which causes snowballed hallucination in future steps. For example, in the below output,
\begin{quote}
    [....previous steps omitted]
    
    Step 3: From city E, we have three options: a flight to city N, a flight to city B, or a flight to city C.

    Step 4: The only option that could potentially lead us to city M is the flight from city E to city C.
    
    [....rest of the output omitted]
\end{quote}
ChatGPT incorrectly states that there are three options in the step 3 (there are only two), inducing the snowballed hallucination \textit{``or a flight to city C''} (ChatGPT can verify that E $\to$ C is not a valid flight in a separate session).
As shown in \autoref{fig:cot_plot}, GPT-4 still has a high overall snowballed hallucination rate at 94.90\% averaged across tasks, and ChatGPT also obtains a similarly high snowballed hallucination rate.

Finally, while our experiments have focused on simple multi-step problems that are suitable for breaking down step-by-step, we hypothesize that hallucination snowballing appears in open-ended text generation more broadly, where one mistake in the generation triggers more \cite{arora-etal-2022-exposure}. 
In these cases, better prompting would neither be able to anticipate nor fix these mistakes.

\subsection{Algorithmic Corrections}\label{subsec:algorithmic_corrections}

\paragraph{Increasing the temperature}
During decoding, the temperature $t$ controls the sharpness of the output distribution, with higher $t$ spreading probability mass away from the model's most likely prediction for each next word.  
Our experiments in \autoref{sec:exp} used greedy decoding, which is equivalent to $t=0$. At $t=0.6$ and $t=0.9$, both error rates and snowballed hallucination rate remain similarly high, in both GPT-4 and ChatGPT (\autoref{fig:temperature_plot}).%

\paragraph{Top-k and nucleus sampling} Using sampling methods such as top-$k$ sampling or nucleus sampling~\cite{holtzman-etal-2020-curious} would not help since they only \emph{narrow} the range of tokens to be considered, and thus can only \emph{increase} the probability that the model will immediately commit to an answer.%

\paragraph{Beam search} The argument for hallucination snowballs in \autoref{sec:why} relies on the fact that, once a model generates some tokens committing to an answer, they remain in the context and influence later generations. One potential way around this is \textit{beam search}, i.e., maintaining a beam of high-probability sequences at each timestep rather than a single sequence. In principle, if some sequences in the beam after the initial token do not commit to an answer (or commit to the right answer), their continuations may eventually have higher probability than those that initially commit incorrectly and later produce incorrect reasoning as a result. If so, beam search would solve the snowball hallucination problem.
Unfortunately, we cannot test the effect of beam search on hallucination snowballs because the OpenAI API does not support beam search.

\paragraph{Learning strategies}
A more general way to further reduce snowballing  might be to change aspects of the pretraining or instruction tuning phases. In particular, a greater emphasis on having the model produce a reasoning chain before generating an answer could be a good way to accommodate its computational limitations and avoid committing to wrong answers that force hallucinations. 

In addition, we hypothesize that finetuning on data with backtracking might improve a model's performance on the tasks we present. This could be accomplished by, for example, giving a question, followed by a wrong solution, and then issuing a phrase like \textit{``Sorry, that was incorrect'' }
before giving the correct solution. This solution is related to the \textit{``Review your previous answer and find problems with your answer.''} prompt from~\citet{computertasks}.

\section{Related Work}

\paragraph{Hallucinations} Hallucination in text generation is a well-studied problem \cite[i.a.]{ rohrbach-etal-2018-object, maynez-etal-2020-faithfulness, raunak-etal-2021-curious} that has recently become more prominent due to ChatGPT's tendency to produce plausible-sounding falsehoods.
Hallucinations are often attributed to knowledge gaps in LMs \cite{zheng2023does}, and several works have shown the promise of using retrieval over knowledge bases to mitigate them \cite{lewis-etal-2020-retrieval,shuster-etal-2021-retrieval-augmentation,peng2023check}.
Our work demonstrates hallucination can be induced from context, thus motivating further mitigation techniques.

Hallucination snowballing is likely the result of \textit{exposure bias}: LMs were only exposed to gold history during training, but during inference, conditions on possibly erroneous previous predictions.
Prior work linked this to compounding hallucinations in machine translation \cite{wang-sennrich-2020-exposure} and open-ended text generation \cite{arora-etal-2022-exposure}.
We go beyond demonstrating error propagation by showing that the propagated errors (which we call snowballed hallucinations) are recognized by the LM itself.%

Our observations are related to previous findings that LMs hallucinate when given questions that contain false presuppositions~\cite[e.g., ``\textit{Which linguist invented the lightbulb?}'';][]{Kim2021WhichLI,Kim2022QA2QA}  or that are otherwise misleading~\cite[e.g., ``\textit{Who really caused 9/11?}'';][]{lin-etal-2022-truthfulqa}, in that faulty context misguides the LM.
However, our work differs in that our questions are not intentionally misleading, showing that this failure mode may be triggered even on innocent information-seeking queries to the LM.

\paragraph{LM (in)consistency}
Our work adds to a growing body of work demonstrating the extent to which LMs are inconsistent across different prompts on the same issue.
For instance, allowing an LM to generate intermediate steps \cite{nye2021work,wei2023chainofthought,  press2022measuring} enables it to reach a different answer than it otherwise would.
Other work has shown that simply prepending ``\textit{Professor Smith was given the following instructions}'' to a prompt can improve performance, despite providing no valuable information about the problem itself \cite{lin-etal-2022-truthfulqa}.

\section{Conclusion}
We define the phenomenon of hallucination snowballing and demonstrate its prevalence in generations from state-of-the-art models, leading to hallucinations on simple facts that wouldn't otherwise occur.
Our findings point to the risk of training language models that prioritize fluency and coherence indiscriminatively at the expense of factuality, and we encourage future work to study remedial actions at all levels of model development.

\section*{Limitations}

We focus on hallucination snowballing in the context of question answering in English, and we do not explore it on other tasks, such as summarization or code generation. 

In addition, we only conduct experiments on two proprietary models, namely ChatGPT and GPT-4, due to their state-of-the-art performance on many benchmarks~\cite{gpt4}. 
Due to the limitations of the APIs for these models, we do not have access to the probability distributions they output and do not have the ability to finetune them. This restricts our ability to explore potential mitigation strategies. Having access to the output distributions would allow us to investigate mitigating the snowballing hallucination issue using alternative sampling methods such as beam search. Having the ability to finetune the model would allow us to explore whether instruction tuning with different annotations could lead to better handling of the questions we use to instigate hallucination snowballing. 

\section*{Acknowledgements}
We thank Sofia Serrano, Yizhong Wang, Yanai Elazar, Michael Hu and Richard Yuanzhe Pang for their valuable feedback and fruitful discussions. While writing this paper, Ofir Press was a visitor at New York University's Center for Data Science, hosted by Kyunghyun Cho.

\bibliography{anthology,custom}
\bibliographystyle{acl_natbib}
\newpage %
\appendix
\section{Dataset Details}
\label{sec:appendix-dataset}
\subsection{Graph Connectivity}\label{subsec:appendix-graph-connectivity}
\begin{figure*}[h!]
    \centering
    \begin{tikzpicture}[scale=0.25]
        \tikzstyle{every node}+=[inner sep=0pt]
        \draw [black] (19.6,-13.5) circle (3);
        \draw (19.6,-13.5) node {$B$};
        \draw [black] (10.8,-22.9) circle (3);
        \draw (10.8,-22.9) node {$F$};
        \draw [black] (27.9,-22.9) circle (3);
        \draw (27.9,-22.9) node {$M$};
        \draw [black] (5.8,-33.2) circle (3);
        \draw (5.8,-33.2) node {$K$};
        \draw [black] (15,-33.2) circle (3);
        \draw (15,-33.2) node {$D$};
        \draw [black] (33,-33.2) circle (3);
        \draw (33,-33.2) node {$I$};
        \draw [black] (54.8,-13.5) circle (3);
        \draw (54.8,-13.5) node {$\mbox{ }N$};
        \draw [black] (46.4,-22.9) circle (3);
        \draw (46.4,-22.9) node {$H$};
        \draw [black] (62.8,-22.9) circle (3);
        \draw (62.8,-22.9) node {$G$};
        \draw [black] (41.8,-33.2) circle (3);
        \draw (41.8,-33.2) node {$A$};
        \draw [black] (50.7,-33.2) circle (3);
        \draw (50.7,-33.2) node {$E$};
        \draw [black] (58.8,-33.2) circle (3);
        \draw (58.8,-33.2) node {$L$};
        \draw [black] (23.5,-33.2) circle (3);
        \draw (23.5,-33.2) node {$J$};
        \draw [black] (68,-33.2) circle (3);
        \draw (68,-33.2) node {$C$};
        \draw [black] (17.55,-15.69) -- (12.85,-20.71);
        \fill [black] (12.85,-20.71) -- (13.76,-20.47) -- (13.03,-19.78);
        \draw [black] (21.59,-15.75) -- (25.91,-20.65);
        \fill [black] (25.91,-20.65) -- (25.76,-19.72) -- (25.01,-20.38);
        \draw [black] (9.49,-25.6) -- (7.11,-30.5);
        \fill [black] (7.11,-30.5) -- (7.91,-30) -- (7.01,-29.56);
        \draw [black] (11.93,-25.68) -- (13.87,-30.42);
        \fill [black] (13.87,-30.42) -- (14.03,-29.49) -- (13.1,-29.87);
        \draw [black] (29.23,-25.59) -- (31.67,-30.51);
        \fill [black] (31.67,-30.51) -- (31.76,-29.57) -- (30.87,-30.02);
        \draw [black] (52.8,-15.74) -- (48.4,-20.66);
        \fill [black] (48.4,-20.66) -- (49.3,-20.4) -- (48.56,-19.73);
        \draw [black] (56.74,-15.78) -- (60.86,-20.62);
        \fill [black] (60.86,-20.62) -- (60.72,-19.68) -- (59.96,-20.33);
        \draw [black] (45.18,-25.64) -- (43.02,-30.46);
        \fill [black] (43.02,-30.46) -- (43.81,-29.93) -- (42.89,-29.53);
        \draw [black] (47.56,-25.67) -- (49.54,-30.43);
        \fill [black] (49.54,-30.43) -- (49.7,-29.5) -- (48.77,-29.89);
        \draw [black] (61.71,-25.7) -- (59.89,-30.4);
        \fill [black] (59.89,-30.4) -- (60.64,-29.84) -- (59.71,-29.48);
        \draw [black] (26.72,-25.66) -- (24.68,-30.44);
        \fill [black] (24.68,-30.44) -- (25.45,-29.9) -- (24.53,-29.51);
        \draw [black] (64.15,-25.58) -- (66.65,-30.52);
        \fill [black] (66.65,-30.52) -- (66.73,-29.58) -- (65.84,-30.03);
    \end{tikzpicture}
    \caption{The underlying graph structure corresponds to example in Table \ref{tab:bfs example}.}
    \label{fig:bfs tree}
\end{figure*}
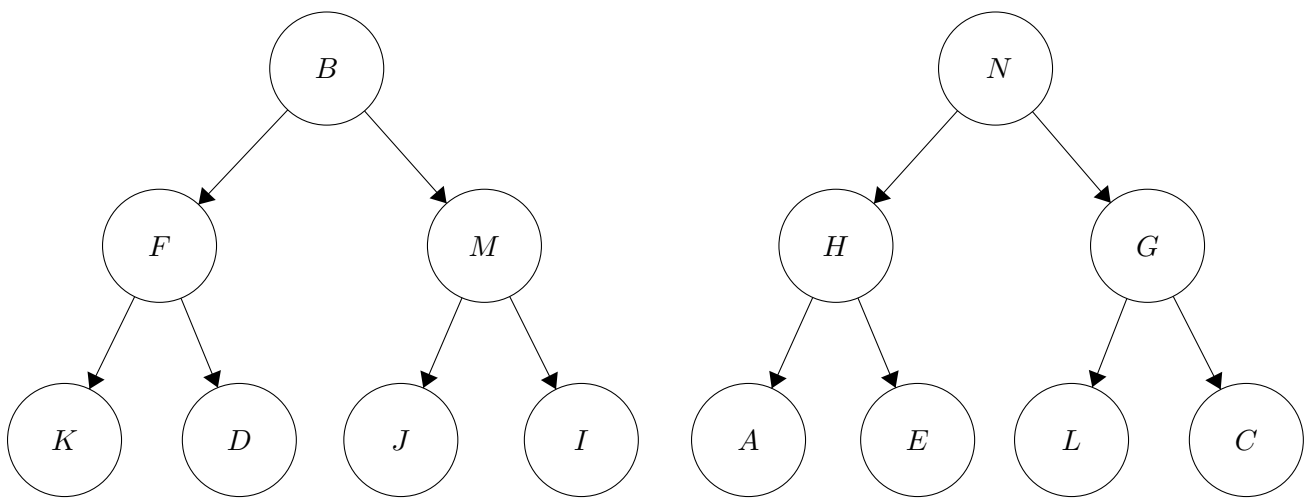

In this dataset, the list of flights can be represented by a directed graph. We generated the flight information to ensure all the graphs share a specific connection pattern, with the node names randomly chosen among the 26 letters in the English alphabet. For an illustration of the underlying graph structure, see \autoref{fig:bfs tree}. %

\subsection{Senator search}
The twelve colleges used in the datasets are: MIT, University of Chicago, Johns Hopkins University, California Institute of Technology, Duke University, Northwestern University, Dartmouth College, Brown University, Vanderbilt University, Rice University, University of Washington. We constructed this list by taking a list of top universities in the U.S. and excluding from it universities which also appeared on The U.S. News \& World Report's list of Top 10 Colleges for Members of Congress. %

\begin{table*}[t!p]
    \centering
    \renewcommand{\arraystretch}{1.2}
    \begin{tabularx}{\textwidth}{X}
        \midrule
        \textbf{Graph Connectivity:}
        Current flight information (the following flights are one-way only, and all the flights available are included below):\\
        There is a flight from city F to city K\\
        There is a flight from city H to city A\\
        There is a flight from city B to city F\\
        There is a flight from city N to city H\\
        There is a flight from city B to city M\\
        There is a flight from city N to city G\\
        There is a flight from city M to city J\\
        There is a flight from city G to city L\\
        There is a flight from city H to city E\\
        There is a flight from city G to city C\\
        There is a flight from city M to city I\\
        There is a flight from city F to city D\\\\
        Question: Is there a series of flights that goes from city B to city E?\\
        \bottomrule 
    \end{tabularx}
    \caption{Input example for the Graph Connectivity task. \autoref{fig:bfs tree} describes the underlying graph structure for this example. We use the same graph structure in all examples, and vary the source and target cities and the labels of the nodes. The cities are literally called ``City A'', ``City B'', and so on.}
    \label{tab:bfs example}
\end{table*}

\begin{table*}[htb]
    \centering
    \begin{tabularx}{\textwidth}{X}
        \midrule
        \textbf{Input}: Current flight information (the following flights are one-way only, and all the flights available are included below):\\
There is a flight from city G to city B\\
There is a flight from city H to city K\\
There is a flight from city L to city M\\
There is a flight from city F to city H\\
There is a flight from city G to city J\\
There is a flight from city B to city I\\
There is a flight from city L to city A\\
There is a flight from city H to city N\\
There is a flight from city B to city D\\
There is a flight from city J to city C\\\\

Question: Is there a series of flights that goes from city F to city I?\\
        \midrule
        \textbf{Output}: Yes, there is a series of flights that goes from city F to city I. The route is as follows:\\\\

        1. City F to City H\\
        2. City H to City K\\
        3. City K to City G\\
        4. City G to City B\\
        5. City B to City I\\
        \midrule
        \textbf{Verification Question}: Current flight information (the following flights are one-way only, and all the flights available are included below):\\
There is a flight from city G to city B\\
There is a flight from city H to city K\\
\textit{[... 10 other rules cut for space ...]}\\\\

Based on the above flight information, is City K to City G a valid flight?\\

        \midrule
         \textbf{Verification Answer}:

        No, based on the above flight information, there is no direct flight from City K to City G.\\
        \bottomrule 
    \end{tabularx}
    \caption{Graph Connectivity dataset verification example, with GPT-4.}
    \label{tab:GC verification example}
\end{table*}

\begin{table*}[htb]
    \centering
    \begin{tabularx}{\textwidth}{X}
        \midrule
        \textbf{Input}: Is 9791 a prime number?\\
        \midrule 
        \textbf{Output}: No, 9791 is not a prime number. It can be factored into 13 × 751.\\
        \midrule
        \textbf{Verification Q1}: Is 9791 divisible by 13? Answer with either Yes or No.\\
        \midrule
        \textbf{Verification A1}: No\\
        \midrule
        \textbf{Verification Q2}: Is 9791 divisible by 751? Answer with either Yes or No.\\
        \midrule
        \textbf{Verification A2}: No\\
        \bottomrule 
    \end{tabularx}
    \caption{Primality Testing dataset verification example, with GPT-4.}
    \label{tab:PT verification example}
\end{table*}

\begin{table*}[htb]
    \centering
    \begin{tabularx}{\textwidth}{X}
        \midrule
        \textbf{Input}: Was there ever a US senator that represented the state of New Hampshire and whose alma mater was University of Pennsylvania?\\
        \midrule 
        \textbf{Output}: Yes, there was a U.S. Senator who represented the state of New Hampshire and whose alma mater was the University of Pennsylvania. His name is John P. Hale, who served as a Senator from 1847 to 1853 and again from 1855 to 1865. He graduated from Bowdoin College in 1827 and later attended the University of Pennsylvania Law School.\\
        \midrule
        \textbf{Verification Q1}: Has John P. Hale represented the state of New Hampshire as US senator?\\
        \midrule
        \textbf{Verification A1}: Yes, John P. Hale represented the state of New Hampshire as a U.S. Senator. He served in the Senate from March 4, 1847, to March 3, 1853, and again from March 4, 1855, to March 3, 1865.\\
        \midrule
        \textbf{Verification Q2}: Was John P. Hale's alma mater University of Pennsylvania?\\
        \midrule
        \textbf{Verification A2}: No, John P. Hale's alma mater was Bowdoin College in Brunswick, Maine.\\
        \bottomrule 
    \end{tabularx}
    \caption{Senator Alma Mater dataset verification example, with GPT-4.}
    \label{tab:SA verification example}
\end{table*}

\section{Additional Results}
We provide the detail breakdown of the question-answering accuracy in \autoref{tab:task performance} and the hallucination detection accuracy in \autoref{tab:verification performance}.

\begin{table*}[hp]
\centering
\begin{tabular}{@{}lllll@{}}
\toprule
&  \multicolumn{3}{c}{Task}& \multirow{2}{*}[0pt]{Average}\\\cmidrule(lr){2-4}
Model    & Graph Connectivity & Primality Testing & Senator Search&  \:\:\: \\ \cmidrule(r){1-1}\cmidrule(l){2-2}\cmidrule(l){3-3}\cmidrule(l){4-4}\cmidrule(l){5-5}
ChatGPT & 410/500 (82.0\%)  & 339/500 (67.8\%) &  153/500 (30.6\%) & 60.13\% \\
GPT-4   & 442/500 (88.4\%)  &  374/500 (74.8\%) &  435/500 (87.0\%) & 83.40\%\\
\bottomrule
\end{tabular}
\caption{Number of \textbf{mistakes} out of the number of samples, the percentage here is the error rate.}
\label{tab:task performance}
\end{table*}

\begin{table*}[hp]
\centering
\begin{tabular}{@{}lllll@{}}
\toprule
&  \multicolumn{3}{c}{Task}& \multirow{2}{*}[0pt]{Average}\\\cmidrule(lr){2-4}
Model    & Graph Connectivity & Primality Testing & Senator Search&  \:\:\: \\ \cmidrule(r){1-1}\cmidrule(l){2-2}\cmidrule(l){3-3}\cmidrule(l){4-4}\cmidrule(l){5-5}
ChatGPT & 396/410 (96.6\%) & 125/339 (36.9\%) &  98/153 (68.6\%) & 67.37\% \\
GPT-4   &  417/442 (94.3\%)  &  346/374 (92.5\%) &  323/435 (74.3\%)  & 87.03\% \\
\bottomrule
\end{tabular}
\caption{Number of snowballed hallucination out of number of hallucination generated in the original output.}
\label{tab:verification performance}
\end{table*}

\begin{table*}[hp]
\centering
\begin{tabular}{@{}lllll@{}}
\toprule
&  \multicolumn{3}{c}{Task}& \multirow{2}{*}[0pt]{Average}\\\cmidrule(lr){2-4}
Model    & Graph Connectivity & Primality Testing & Senator Search&  \:\:\: \\ \cmidrule(r){1-1}\cmidrule(l){2-2}\cmidrule(l){3-3}\cmidrule(l){4-4}\cmidrule(l){5-5}
ChatGPT & 139/500 (27.8\%)  & 2/500 (0.4\%) &  0/500 (0.0\%) & 9.40\% \\
GPT-4   & 21/500 (4.2\%)  &  37/500 (7.4\%) &  0/500 (0.0\%) & 3.87\%\\
\bottomrule
\end{tabular}
\caption{Number of \textbf{mistakes} out of the number of samples, the percentage here is the error rate, using ``Let's think step by step'' prompt.}
\label{tab:lts task performance}
\end{table*}

\begin{table*}[hp]
\centering
\begin{tabular}{@{}lllll@{}}
\toprule
&  \multicolumn{3}{c}{Task}& \multirow{2}{*}[0pt]{Average}\\\cmidrule(lr){2-4}
Model    & Graph Connectivity & Primality Testing & Senator Search&  \:\:\: \\ \cmidrule(r){1-1}\cmidrule(l){2-2}\cmidrule(l){3-3}\cmidrule(l){4-4}\cmidrule(l){5-5}
ChatGPT & 123/139 (88.5\%) & 0/2 (0\%) &  0/0 (N/A) & 44.25\% \\
GPT-4   & 20/21 (95.2\%)  &  35/37 (94.6\%) &  0/0 (N/A)  & 94.90\% \\
\bottomrule
\end{tabular}
\caption{Number of snowballed hallucination out of number of hallucination generated in the original output, using ``Let's think step by step'' prompt.}
\label{tab:lts verification performance}
\end{table*}

\begin{table*}[hbp]
\centering
\begin{tabular}{@{}lllll@{}}
\toprule
Model    & Graph & Prime & Senator & Average\\ \midrule
ChatGPT ($t=0.0$) & 410/500 (82.0\%) & 339/500 (67.8\%) & 153/500 (30.6\%) & 60.13\%\\
ChatGPT ($t=0.6$) & 407/500 (81.4\%)  & 310/500 (63.2\%) & 155/500 (31.0\%) &  58.53\%\\
ChatGPT ($t=0.9$)   & 403/500 (80.6\%)   &  312/500 (62.4\%) &  163/500 (32.6\%)  &  58.53\%\\ \midrule
GPT-4 ($t=0.0$) & 442/500 (88.4\%) & 374/500 (74.8\%) & 435/500 (87.0\%) & 83.40\%\\
GPT-4 ($t=0.6$) & 438/500 (87.6\%) & 365/500 (75.4\%) & 423/500 (84.6\%) & 82.53\%\\
GPT-4 ($t=0.9$) & 437/500 (87.4\%) & 377/500 (73.0\%) & 423/500 (84.6\%) & 81.67\%\\
\bottomrule
\end{tabular}
\caption{Number of \textbf{mistakes} out of the number of samples, the percentage here is the error rate, with different temperature setups}
\label{tab:temperature experiment - accuracy}
\end{table*}

\begin{table*}[hbp]
\centering
\begin{tabular}{@{}lllll@{}}
\toprule
Model    & Graph & Prime & Senator & Average\\ \midrule
ChatGPT ($t=0.0$) & 396/410 (96.6\%) & 125/339 (36.9\%) &  98/153 (68.6\%) & 67.37\% \\
ChatGPT ($t=0.6$) & 396/407 (97.3\%)  & 113/310 (36.5\%) & 103/155 (66.5\%) &  66.77\%\\
ChatGPT ($t=0.9$)   & 399/402 (99.3)\%   &  116/312 (37.2\%) &  104/163 (63.8\%)  &  66.77\%\\ \midrule
GPT-4 ($t=0.0$) &  417/442 (94.3\%)  &  346/374 (92.5\%) &  323/435 (74.3\%)  & 87.03\% \\
GPT-4 ($t=0.6$) & 402/438 (91.8\%) & 324/365 (88.8\%) & 329/423 (77.8\%) & 86.13\%\\
GPT-4 ($t=0.9$) & 398/437 (91.1\%) & 331/377 (87.8\%) & 320/423 (75.7\%) & 84.87\%\\
\bottomrule
\end{tabular}
\caption{Number of \textbf{mistakes} out of the number of samples, the percentage here is the error rate, with different temperature setups}
\label{tab:temperature experiment - snowball}
\end{table*}

\begin{table*}[bht]
    \centering
    \begin{tabularx}{\textwidth}{X}
        \midrule
        \textbf{Input}: Was there ever a US senator that represented the state of North Carolina and whose alma mater was Dartmouth College?\\
        \midrule
        \textbf{Output}: Yes, there was a U.S. Senator who represented the state of North Carolina and whose alma mater was Dartmouth College. His name was Willis Smith, and he served as a Senator from 1950 to 1953.\\
        \midrule
        \textbf{Verification Q1}: Has Willis Smith represented the state of North Carolina as a US senator?\\
        \midrule
        \textbf{Verification A1}: Yes, Willis Smith represented the state of North Carolina as a U.S. Senator. He served from 1950 until his death in 1953.\\
        \midrule
        \textbf{Verification Q2}: Was Willis Smith's alma mater Dartmouth College?\\
        \midrule
        \textbf{Verification A2}: Yes, Willis Smith's alma mater was Dartmouth College. He graduated from there in 1917.\\
        \bottomrule 
    \end{tabularx}
    \caption{An example of GPT-4 failing the verification test for Senator Search task. Willis Smith was a senator from North Carolina, but he has never attended Dartmouth College.}
    \label{tab:failed verification}
\end{table*}

\end{document}